\documentclass[research]{f}
\usepackage{tcolorbox}
\usepackage{multicol}
\usepackage{multirow}
\usepackage{bm}
\usepackage{arydshln}
\usepackage{xcolor}
\usepackage[normalem]{ulem}

\title{Exploiting User Comments for Early Detection of Fake News Prior to Users' Commenting}
\author[1,2]{Qiong Nan}
\author[1]{Qiang Sheng}
\author*[1,2]{Juan Cao}
\author[1]{Yongchun Zhu}
\author[1]{Danding Wang}
\author[3]{Guang Yang}
\author[1]{Jintao Li}
\address[1]{Institute of Computing Technology, Chinese Academy of Sciences, Beijing, 100190, China}
\address[2]{University of Chinese Academy of Sciences, Beijing, 100049, China}
\address[3]{Zhongguancun Laboratory, Beijing, 100080, China}
\fcssetup{
  corr-email     = {caojuan@ict.ac.cn},
}
\begin{abstract}
Both accuracy and timeliness are key factors in detecting fake news on social media. 
However, most existing methods encounter an accuracy-timeliness dilemma: 
Content-only methods guarantee timeliness but perform moderately because of limited available information, while social con-text-based ones generally perform better but inevit-ably lead to latency because of social context accumulation needs. 
To break such a dilemma, a feasible but not well-studied solution is to leverage social contexts (e.g., comments) from historical news for training a detection model and apply it to newly emerging news without social contexts. This requires the model to (1) sufficiently learn helpful knowledge from social contexts, and (2) be well compatible with situations that social contexts are available or not. 
To achieve this goal, we propose to absorb and parameterize useful knowledge from comments in historical news and then inject it into a content-only detection model.
Specifically, we design the \underline{C}omments \underline{AS}sisted \underline{F}ak\underline{E} \underline{N}ews \underline{D}etection method (CAS-FEND), which transfers useful knowledge from a comment-aware teacher model to a content-only student model and detects newly emerging news with the student model. 
Experiments show that the CAS-FEND student model outperforms all content-only methods and even comment-aware ones with 1/4 comments as inputs, demonstrating its superiority for early detection.
\end{abstract}
\keywords{fake news detection, knowledge distillation, early detection}

\begin{document}
\section{Introduction}
\label{sec1}
Fake news on online social media spreads rapidly and often causes severe social impacts in a short time.
For example, a claim that 5G networks boost the spread of the coronavirus circulated quickly and widely in communities during the COVID-19 pandemic, which led to many destructive acts, including the burning of 5G base stations and attacks on related facilities, affecting the stability of communication infrastructure~\cite{5GFakeNews}. 
To minimize the potential harm, it is imperative to detect newly emerging fake news as early as possible. 

In the existing literature, detecting fake news at the early stage is still challenging. Most fake news detection methods focus on capturing the difference in the elicited social contexts between fake and real news, such as information propagation patterns~\cite{Fang_Unsupervised}, user comments~\cite{Shu_dEFEND, dualemotion}, and engaged user profiles~\cite{Castillo_Information, Shu_Beyond, Shu_MIPR2018_Understanding} and achieve good detection accuracy.
Though accurate, these methods could inevitably lead to detection latency because social context information requires time to accumulate.
To reduce the responsive time and perform instant detection, other researchers explore the possibility of detecting fake news with only news content as inputs, mostly mining information based on stylometric kn-owledge~\cite{ potthast2017stylometric, Przybyla_Capturing} or learning text representation using neural networks~\cite{eann, FakeBERT_Kaliyar}. Though they can operate even with no social contexts available, their performance is generally inferior to social context-based ones due to limited available information.
This comparison reveals that existing methods encounter an \textbf{accuracy-timeliness dilemma} that the detection performance of comment-aware methods\footnote{ We focus on comment-aware methods among those using social contexts because user comments include diverse clues based on crowd wisdom and are easy to collect. For example, Shu et al.~\cite{Shu_ECMLPKDD2020_WeakSupervision} leverage user comments to provide weak labels for unlabeled data in fake news detection.} increases as more comments are available and the performance gap between content-only and comm-ent-aware methods is as large as 0.11 in macro F1 (experimentally exposed in Tables~\ref{tab:results_stu}, \ref{tab:results_tea} and~\ref{tab: comment_num_test} latter).
An urgent need emerges: How to break such a dilemma and find a way to detect newly emerging fake news as early as possible while guaranteeing performance in real-world applications?

\begin{figure}
    \centering
\includegraphics[width=1.0\linewidth]{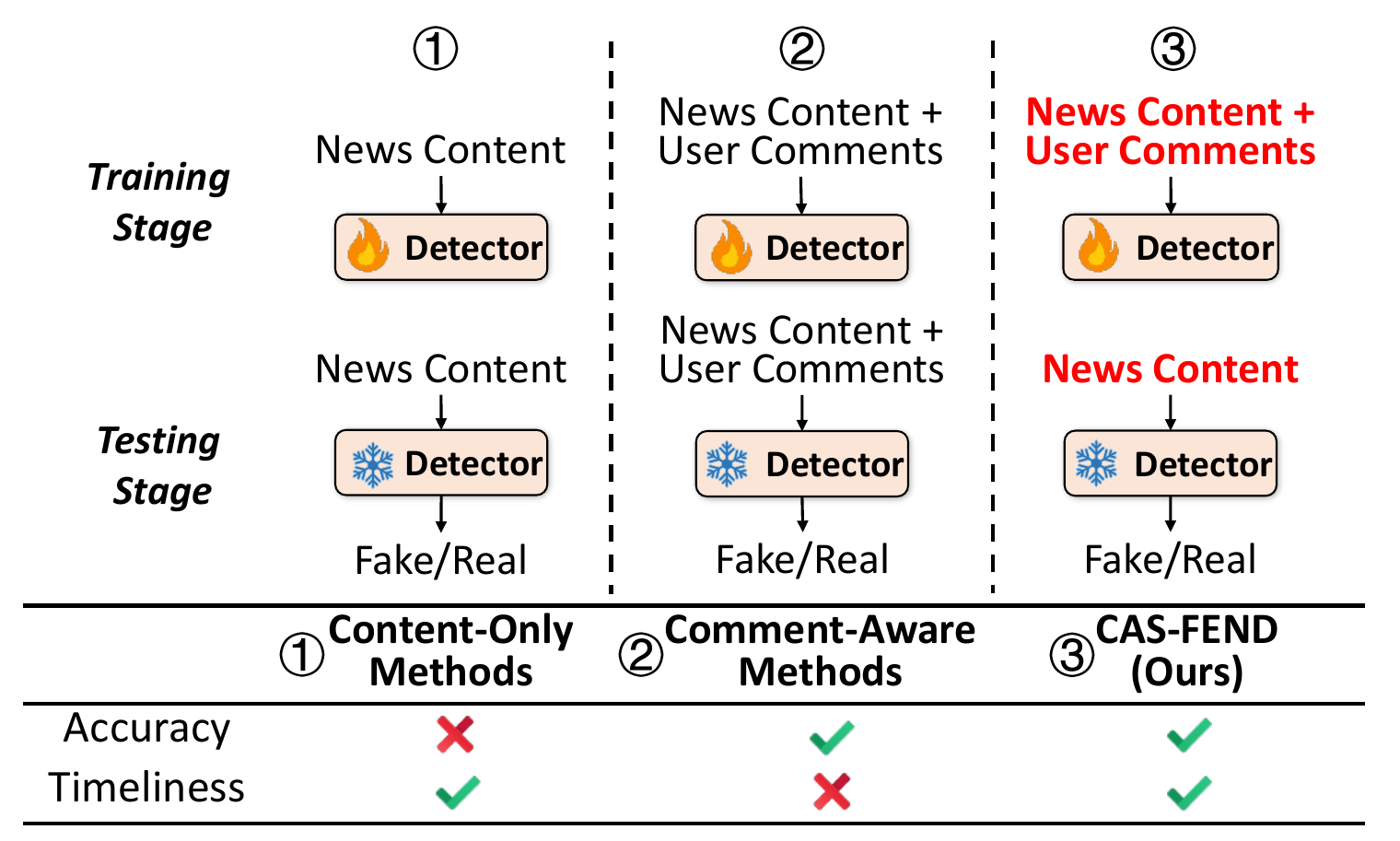}
    \caption{Difference between existing fake news detection methods (content-only and comment-aware) and ours in training-testing settings.}
    \vspace{-0.5cm}
    \label{fig:difference}
\end{figure}

In this paper, we explore how to maintain accuracy and timeliness for fake news detection by reducing the performance gap between content-only and comment-aware methods.
Essentially, the performance gap between the two types of methods is derived from the information gap: The worse-performing content-only methods could use no information from user comments as comment-aware ones do for the sake of timeliness. 
This suggests that a feasible solution is to find a supplementary information source for content-only methods that could serve as a surrogate for the \textit{possibly existing comments} in the future, which is currently understudied. Intuitively, the surrogate would enable the content-only methods to have an ``educated guess'' of additional information in future user comments, thus deepening their understanding of the news content itself. 
As an instantiation, we propose exploiting user comments \textit{in the historical news} as a surrogate to assist the training of content-only fake news detectors. \figurename~\ref{fig:difference} illustrates the difference in training-testing settings between our solution and existing ones.

To this end, we design the \underline{C}omments \underline{AS}sisted \underline{F}ak\underline{E} \underline{N}ews \underline{D}etection method (CAS-FEND), which absorbs and parameterizes useful knowledge from comments in historical news and then injects it into the content-only fake news detection model.
CAS-FEND is implemented with a teacher-student framework, where the comment-aware teacher model is trained with both news content and user comments, and the content-only student model is subsequently trained with news content and guided by the teacher model.
In practice, the student model is used to detect newly emerging fake news.

For the teacher model training, we aim to absorb and parameterize useful comment knowledge from semantic and emotional perspectives.
To obtain semantic-level knowledge, we apply the co-attention mechanism between news content and user comments to enhance the understanding of both sides.
To obtain emotional-level knowledge, we extract numerical social emotions from user comments which reflect the topics and tones of the news and transform them into continuous features. 
We aggregate semantic features of news content and user comments after the co-attention mechanism and the continuous social emotion feature to get the overall feature of the teacher model for final classification.
For the student model training, we aim to transfer useful knowledge from the teacher model.
For the feasibility and convenience of knowledge transferring, we construct a homologous structure for the student model as the teacher model. 
Specifically, we get the content's semantic feature by assigning learnable weights to different tokens in news content through the mask-attention mechanism and acquire the possible elicited social emotion feature (\textit{i.e.}, virtual social emotion) via a Social Emotion Predictor with news content as input. 
We aggregate the content's semantic feature and the virtual social emotion feature to get the overall feature of the student model for final classification.
To transfer useful knowledge from the teacher model to the student model, we adaptively distill knowledge from semantic, emotional, and overall perspectives. 

Our contributions are summarized as follows:

\textbf{(1) Important observation:} We investigate the problem of guaranteeing both timeliness and accuracy for fake news detection and point out that the performance gap is derived from the information gap between detection methods with and without comments.

\textbf{(2) Practical solution:} We design CAS-FEND, a novel method capable of producing a surrogate for possibly existing comments to obtain high performance for early detection of fake news.

\textbf{(3) Superior performance:} The student model of CAS-FEND outperforms all content-only methods and even comment-aware ones with a quarter of comments, and obtains the best performance with highly-skewed data.

\section{Related Work}
\textbf{Fake News Detection.} Existing works exploit various information to detect fake news, and they can be grouped into two clusters: content-only and social-context-based fake news detection methods~\cite{FakeNewsSurvey_Shu, Guo_FakeNews_FCS}.
For content-only fake news detection, researchers extract useful features from the raw news content, which can achieve immediate fake news detection (timeliness). 
Specifically, earlier works incorporate a wide variety of manually crafted features by calculating specific words' frequency, tagging part of speech, etc~\cite{Castillo_Information}, others detect fake news based on writing styles~\cite{ potthast2017stylometric, Przybyla_Capturing}. 
With the development of deep learning, many researchers exploit recurrent neural networks (RNNs) or convolutional neural networks (CNNs) to extract high-level features of news content for fake news detection~\cite{Ma_RNN, Hussain_CNNFusion}.
Kaliyar et al.~\cite{FakeBERT_Kaliyar} combine the BERT encoder with different parallel blocks of the single-layer CNNs for fake news detection.
To deal with domain and temporal distribution shifts, researchers exploit various strategies to enhance the generalization ability of detectors, such as adversarial training~\cite{eann}, transfer learning~\cite{Silva_EDDFN, Nan_DITFEND, Huang_MetaPrompt}, contrastive learning~\cite{Lin_LowResource}, and entity bias mitigating~\cite{endef}. 
Besides, Sheng et al.~\cite{Sheng_ZoomOut, Sheng_Prefend} exploit external knowledge to enhance fake news detection performance.
There are also works investigating multi-modal fake news detection~\cite{Qi_Entity, Wu_MultiReadingHabits, Sun_TKDE2023_Inconsistent, Hu_TKDE2023_Causal, Hu_mutualKD, buyuyan}. 
In this paper, we focus on the internal information of the textual modality.

The content-only methods neglect the important role of social context information, which arises thr-ough the diffusion of news content and often contains helpful crowd feedback.
Therefore, multiple types of social context information have been explored for fake news detection, such as user attributes~\cite{Liu_FNED, Shu_Beyond, Nguyen_CIKM2020_FANG}, propagation structures~\cite{Lu2020_GCAN, Fang_Unsupervised, Wang_DHCF, Gao_Rumor_FCS}, user comments~\cite{dualemotion, Shu_dEFEND, Nan_GenFEND, Wan_DELL}, etc. 
Among these types of information, user comments are the most widely used for they include diverse clues based on crowd wisdom and are easy to collect~\cite{Wei_context, HUANG_legal, Wang_powercomments}.
Shu et al.~\cite{Shu_ECMLPKDD2020_WeakSupervision} leveraged signals from social engagements to generate weak labels for early fake news detection, which demonstrates the importance of user comments as well as the strong correlations with news content.
Many researchers enhance feature representations of news content and user comments by performing interactions between them. From the semantic perspective, Shu et al.~\cite{Shu_dEFEND} utilized a co-attention mechanism to capture useful features from news content and user comments jointly.
From the emotional perspective, Zhang et al.\cite{dualemotion} proved the effectiveness of the social emotion feature to enhance fake news detection performance. 
However, the social-context information such as user comments takes time to accumulate and inevitably leads to detection latency.

\textbf{Knowledge Distillation.} The main goal of knowledge distillation (KD) is to transfer knowledge from a superior teacher model to a student model~\cite{LI_KDCrowd, JI_TeaCoop}. 
Based on the types of distilled knowledge, it can be grouped into three categories: response-based, feature-based, and relation-based KD~\cite{gou2021_kdsurvey}. Response-based KD usually optimizes the student's output with the teacher's final output~\cite{hinton2015distilling, Zhou_Rethinking, LI_KDCrowd}. 
It can be easily applied to multiple tasks but ignores the valuable information encoded in the hidden layers of the neural network, which limits the student model's performance.
As an improvement, feature-based KD transfers more meaningful information from hidden layers to the student model~\cite{romero2014fitnets, Komodakis2017_Paying, Zhang_MultiGranularity}. 
Relation-based KD aims to exploit cross-sample relationships across the wh-ole dataset or cross-layer interactive information as meaningful knowledge, which is high-ordered and more complicated. The former usually exploits a relational graph to capture structure dependencies~\cite{Chen_Wang_Zhang_2018, Tian2019_Contrastive, LI_KDCrowd}, while the latter encodes the information flowing procedure~\cite{ Passalis_2020_CVPR, pmlr-v97-jang19b}. Therefore, relation-based KD is more suitable for scenarios where there exists an obvious correlation among datasets, such as user-item interactions in recommendation.
In this paper, we take advantage of the feature-based KD strategy to transfer knowledge from the comment-aware teacher model to the content-only student model.

\section{Problem Formulation}
We assume that a piece of social media news $O$ is a ``content-comments'' pair, which includes the news content $P$ posted by the publisher user and the elicited comments $C = \{c^1, c^2, \ldots, c^{|C|} \}$ from reader users, where $|C|$ is the number of comments.
Fake news detection is typically formulated as a binary classification task between fake and real news~\cite{FakeNewsSurvey_Shu, Li_TIST2022_Dynamic}. Formally, given $O$, we aim to learn a model $\text{f}(O)\rightarrow y$, where $y$ is the ground-truth veracity label with $1$ denoting fake and $0$ denoting real. If $\text{f}$ is a content-only model, $C$ will be an empty set by default.
In this work, we also utilize social emotion $e$ from comments as auxiliary information for training the teacher model. 

\section{Method}
We aim at the scenario where only news content is available for newly published news pieces, but both news content and comments can be obtained for massive historical data. Our goal is to perform fake news detection for these newly published news pieces with the help of historical data, especially the accumulated user comments.
The design of the CAS-FEND method consists of two-stage training (teacher model training and student model training), which mainly focuses on the following technical challenges: (1) How to absorb useful knowledge from comments during the teacher model training; (2) How to transfer useful knowledge from the teacher model during the student model training. 

\begin{figure*}[t]
\centering
\includegraphics[width=\textwidth]{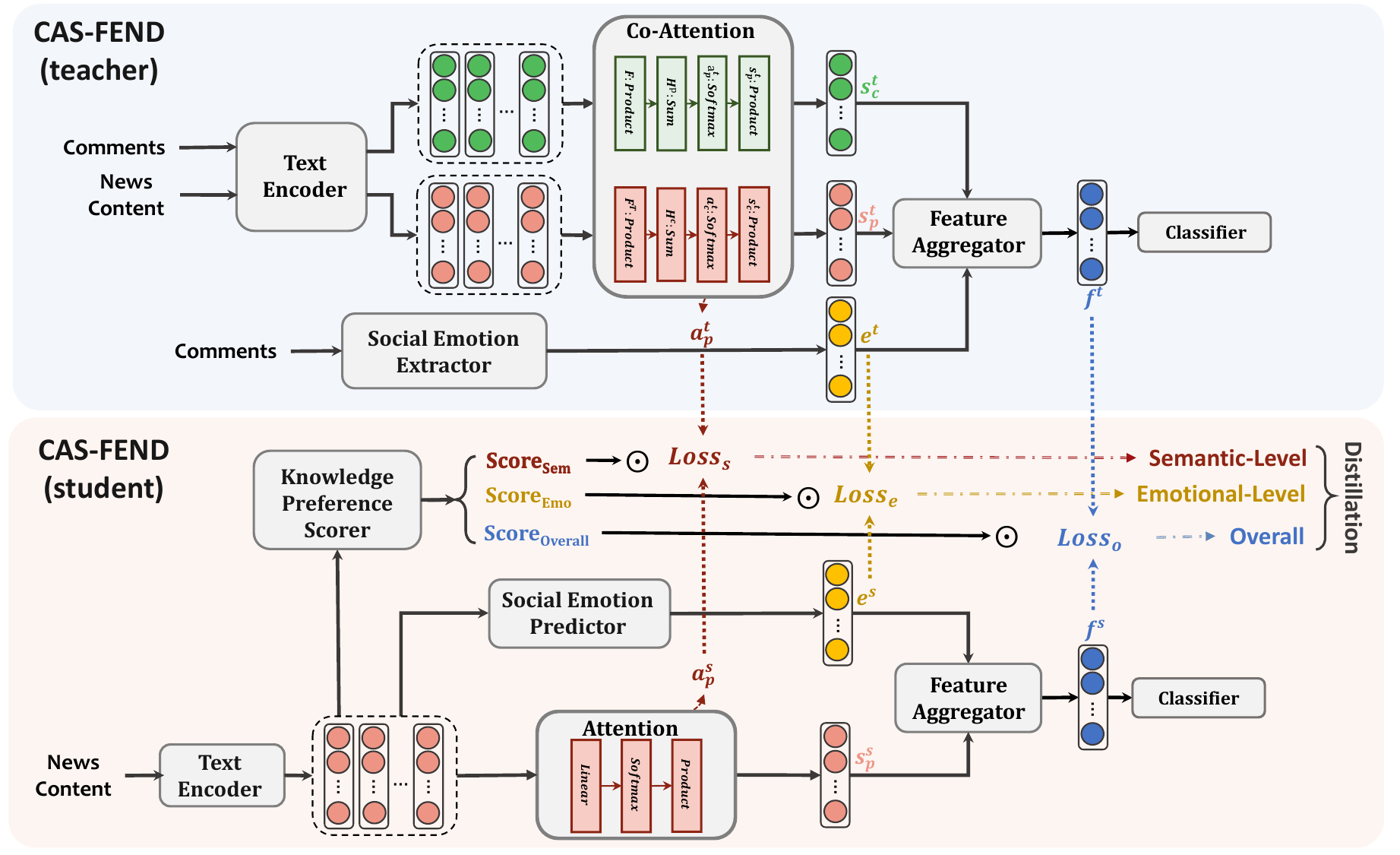}
\caption{Overall architecture of the \textbf{Comments Assisted Fake News Detection (CAS-FEND) framework}, which consists of a \textbf{teacher model} and a \textbf{student model}. During the teacher model training, we exploit both news content and comments by aggregating the semantic feature of news content ($s^t_p$) and user comments ($s^t_c$) and the social emotion feature $e^t$. When training the student model, we only input the content and freeze the well-trained teacher model as guidance. We adopt a \textbf{Social Emotion Predictor} to get an alternative \textit{virtual} social emotion feature $e^s$ and aggregate it with content semantic feature $s^s_p$. With the help of \textbf{Knowledge Preference Scorer}, the student feature learning is guided by the teacher model through a loss-based adaptive knowledge distillation at the semantic, emotional, and overall knowledge levels.}
\label{fig:model}
\end{figure*}

The overall architecture of CAS-FEND is shown in \figurename~\ref{fig:model}. The core idea of our method is to train a content-only student model for accurate and instant fake news detection by distilling useful knowledge from a comment-aware teacher model.  
In general, we adaptively distill knowledge from semantic, emotional, and overall perspectives.

\subsection{Teacher Model Training}\label{Section: teacher}
For the subsequent knowledge absorption from accumulated user comments, the teacher model takes both news content and user comments of historical news pieces as input. 

Usually, from the semantic perspective, the core information of a piece of news only comes from one important sentence or several words, and other auxiliary information contributes less to judging the news' authenticity.
Similarly, user comments may also be noisy in explaining why the corresponding news piece is fake. 
Paying more attention to the key information in news content and user comments is beneficial to detecting fake news. 
We exploit the token-level co-attention mechanism~\cite{CoAttention} between news content and user comments to select key information for both sides because they are related inherently to provide important cues for each other and the attention mechanism has been widely used in natural language processing tasks to assign higher weights to important tokens.
Let $d$ denote the encoding dimension, we tokenize and encode each piece of news as $\boldsymbol{P}= [ \boldsymbol{p}_1, \boldsymbol{p}_2, \ldots, \boldsymbol{p}_M ] \in \mathbb{R}^{d \times M}$, where $M$ is the number of tokens of news content.
We string the corresponding user comments in chronological order as $C_{all} = c_1 \bigoplus c_2 \ldots \bigoplus c_{\left | C \right |}$ ($\bigoplus$ is the concatenating operator), 
tokenize and encode $C_{all}$ as $\boldsymbol{C} =  [ \boldsymbol{c}_1, \boldsymbol{c}_2, \ldots, \boldsymbol{c}_N ] \in \mathbb{R}^{d \times N}$, where $N$ is the number of tokens of concatenated comments $C_{all}$.
Following Lu et al.~\cite{CoAttention}, we first compute the affinity matrix $\boldsymbol{F} \in \mathbb{R}^{M \times N}$ as: 
\begin{equation}
    \boldsymbol{F} = \tanh{(\boldsymbol{P}^\top \boldsymbol{W}_l\boldsymbol{C})},
\end{equation}
where $\boldsymbol{W}_l \in \mathbb{R}^{d \times d}$ is a learnable weight matrix. Similar to Lu et al.~\cite{CoAttention}, we consider this affinity matrix as a feature and learn to predict the content and comments attention maps respectively as follows:
\begin{equation}
    \begin{aligned}
        \boldsymbol{H}_p &=\tanh{(\boldsymbol{W}_p\boldsymbol{P} + (\boldsymbol{W}_c\boldsymbol{C})\boldsymbol{F}^\top)}, \\
        \boldsymbol{a}_p^t &= \text{softmax}(\boldsymbol{w}^\top_{hp}\boldsymbol{H}_p), \\
        \boldsymbol{H}_c &= \tanh{(\boldsymbol{W}_c\boldsymbol{C} + (\boldsymbol{W}_p\boldsymbol{P})\boldsymbol{F})}, \\
        \boldsymbol{a}_c^t &= \text{softmax}(\boldsymbol{w}^\top_{hc}\boldsymbol{H}_c),
    \end{aligned}
\end{equation}
where $\boldsymbol{W}_p, \boldsymbol{W}_c \in \mathbb{R}^{k \times d}, \boldsymbol{w}^\top_{hp}, \boldsymbol{w}^\top_{hc} \in \mathbb{R}^k$ are weight parameters, $k$ is the hidden dimension of the co-attention mechanism, $\boldsymbol{a}_p^t \in \mathbb{R}^M, \boldsymbol{a}_c^t \in \mathbb{R}^N$ are the attention weight distributions of tokens in news content and comments respectively. 
Based on the attention weights $\boldsymbol{a}_p^t$ and $\boldsymbol{a}_c^t$, the feature of the news content and comments are calculated as the weight-ed sum of the token-level encoding features:
\begin{equation}
    \boldsymbol{s}_p^t = \sum_{i = 1}^M {a_p^t}_i\boldsymbol{p}_i, \ 
    \boldsymbol{s}_c^t = \sum_{i = 1}^N {a_c^t}_i\boldsymbol{c}_i,
\end{equation}
where $\boldsymbol{s}_p^t, \boldsymbol{s}_c^t \in \mathbb{R}^{d}$ are semantic features for news content and comments after the co-attention mechanism. 
As the attention weights $\boldsymbol{a}_p^t$ on news content are obtained through interacting with user comments, it reflects the understanding of content with the help of the semantic knowledge in comments. 

Besides, existing research has shown that fake news often evokes high-arousal or uprising emotions from the crowd to spread more virally, and the emotion feature from comments (\textit{i.e.}, social emotion) can enhance news understanding and improve fake news detection~\cite{dualemotion}.  
Therefore, we leverage user comments from the emotional perspective and extract the numerical emotion feature $\boldsymbol{e}_{original}^t \in \mathbb{R}^{d_{emo}}$ from comments (\textit{i.e.}, original social emotion) based on several widely-used resources, including VAD-ER~\cite{bird2009natural}, NRC Emotional Dictionary~\cite{mohammadword}, and How-Net~\cite{dong2003hownet}. 
Specifically, the numerical emotion feature covers eight types: Emotional Lexicon, Emotional Intensity, Sentiment Score, Emoticons Count, Punctuations Count, Sentiment Words Count, Personal Pronoun, and Upper Case Count (only applicable for English)\footnote{The details of these emotion features can be found in~\cite{dualemotion}.}.

The original social emotion $\boldsymbol{e}_{original}^t$ is encoded as a dense embedding $\boldsymbol{e}^t \in \mathbb{R}^{d}$ through a multi-layer perceptron, which is the same dimension as the comments' semantic feature $\boldsymbol{s}_c^t$.
We aggregate the news content feature $\boldsymbol{s}_p^t$, the user comment feature $\boldsymbol{s}_c^t$ and social emotion feature $\boldsymbol{e}^t$ through the Feature Aggregator, which sums the three features with different weighting values to obtain the overall feature $\boldsymbol{f}^t$, which is:
\begin{equation}
    \boldsymbol{f}^t = w_p^t \cdot \boldsymbol{s}_p^t + w_c^t \cdot \boldsymbol{s}_c^t + w_e^t \cdot \boldsymbol{e}^t,
\end{equation}
where $0 \le w_p^t, w_c^t, w_e^t \le 1$ are learnable parameters, and $\boldsymbol{f}^{t} \in \mathbb{R}^{d}$ is the overall feature in the teacher model for the final classification:
\begin{equation}
    \hat{y} = \text{MLP}(\boldsymbol{f}^{t}).
\end{equation}
The optimization goal of the teacher model training is to minimize the cross-entropy loss:
\begin{equation}
    \mathcal{L}_{tea}(\theta_{tea}) = -y\log \hat{y} - (1-y)\log (1-\hat{y}),
\end{equation}
where $\theta_{tea}$ are parameters of the teacher model, $y$ is the ground-truth label, and $\hat{y}$ is the predicted label.

\subsection{Student Model Training}
During the student model training, the parameters of the teacher model are frozen and will not be updated. 
We aim to transfer knowledge from the trained-over-powerful teacher model to the content-only student model to enhance its early fake news detection ability. 
Based on the comments utilization in the teacher model, we consider knowledge transferring from semantic, emotional, and overall perspectives.
\subsubsection{Semantic-Level Knowledge Distillation}

We transfer semantic-level knowledge from user comments by learning the relative importance of different parts of news content from the teacher mo-del, which is reflected by weights obtained through the co-attention mechanism.
For the student model, we assign learnable attention weights $\boldsymbol{a}_p^s \in \mathbb{R}^M$ to all tokens, and obtain the content semantic feature as follows:
\begin{equation}
    \boldsymbol{s}_p^s = \sum_{i = 1}^M {a_p^s}_i\boldsymbol{p_i}.
\end{equation}
We optimize the learnable attention weights $\boldsymbol{a}_p^s$ by referring to $\boldsymbol{a}_p^t$ and minimizing the Mean Square Error (MSE) loss as follows:
\begin{equation}
    \mathcal{L}_{s}(\theta_{stu}) = \frac{1}{\left |p\right |}\sum_{i = 1}^{M} ({a_p^t}_i - {a_p^s}_i)^2,
\end{equation}
where $\theta_{stu}$ are parameters of the student model.
\subsubsection{Emotional-Level Knowledge Distillation}

Based on the tight correlation between news content and aroused emotion in user comments, we calculate the virtual social emotion $\boldsymbol{e}^s$ through the Social Emotion Predictor.
To protect coarse-grained information of news content, we input the token-level embeddings to the Social Emotion Predictor but not the semantic feature $\boldsymbol{s}_p^s$.
Specifically, the Social Emotion Predictor consists of a mask attention module (MaskAttention) to aggregate token embeddings and a multi-layer perception (MLP) for feature mapping, which is:
\begin{equation}
    \hat{\boldsymbol{e}}^s = \text{MaskAttention}(\boldsymbol{P}), \quad
    \boldsymbol{e}^s = \text{MLP}(\hat{\boldsymbol{e}}^s),
\end{equation}
where $\hat{\boldsymbol{e}}^s$ is the temporary aggregated feature and $\boldsymbol{e}^s \in \mathbb{R}^d$ is the obtained virtual social emotion feature. 
The virtual social emotion feature $\boldsymbol{e}^s$ is optimized by referring to  $\boldsymbol{e}^t$ and minimizing the Mean Square Error (MSE) loss as follows:
\begin{equation}
    \mathcal{L}_{e}(\theta_{stu}) = \frac{1}{d}\sum_{i = 1}^{d} (e^t_i - e^s_i)^2.
\end{equation}
\subsubsection{Overall Knowledge Distillation}

The superior ability of the teacher model compared to the student model has a direct relationship with
the overall feature $\boldsymbol{f}^{t}$ for classification. 
Therefore, we transfer the knowledge of the overall feature to improve the student model's detection performance.
Specifically, the overall feature in the student model $\boldsymbol{f}^s \in \mathbb{R}^{d}$ is obtained by aggregating the content feature $\boldsymbol{s}_p^s$ and the virtual social emotion feature $\boldsymbol{e}^s$, which is:
\begin{equation}
    \boldsymbol{f}^{s} = w_p^s \cdot \boldsymbol{s}_p^s + w_e^s \cdot \boldsymbol{e}^s,
\end{equation}
where $0 \le w_p^s, w_e^s \le 1$ are learnable parameters and $\boldsymbol{f}^{s} \in \mathbb{R}^{d}$ is the overall feature in the student model for the final classification:
\begin{equation}
    \hat{y} = \text{MLP}(\boldsymbol{f}^{s}).
\end{equation}
The overall feature $\boldsymbol{f}^s$ is optimized by gradually approaching $\boldsymbol{f}^{t}$ via minimizing the MSE loss:
\begin{equation}
    \mathcal{L}_{o}(\theta_{stu}) = \frac{1}{d}\sum_{i = 1}^{d} (f^t_i - f^s_i)^2,
\end{equation}
where $\theta_{stu}$ are parameters of the student model. 
\subsubsection{Adaptive Fusion}

To utilize semantic, emotional, and overall knowledge effectively, we adjust the importance of three types of knowledge adaptively based on the news content via a Knowledge Preference Scorer, which consists of a mask attention module (MaskAttention), and a multi-layer perceptron (MLP) followed by a Softmax function.
The Knowledge Preference Scorer takes the token-level embeddings of the news content as input and outputs three corresponding scores as weighted values of three types of distilled knowledge, which is:
\begin{equation}
    \begin{aligned}
        \boldsymbol{e}^{scr} &= \text{MaskAttention}(\boldsymbol{P}), \\
    \boldsymbol{scr} &= \text{Softmax}(\text{MLP}(\boldsymbol{e}^{scr})), 
    \end{aligned}
\end{equation}
where $\boldsymbol{scr} = \left [scr_{o}, scr_{s}, scr_{e}\right ]$ is a 3-dimensional vector which indicates weights of the corresponding knowledge. 

The student model is optimized by minimizing the cross-entropy loss $\mathcal{L}_{cls}(\theta_{stu})$ for binary classification and aggregated MSE loss $ \mathcal{L}_{distill}(\theta_{stu})$ for distillation as follows:
\begin{equation}
    \begin{aligned}
        \mathcal{L}_{distill}(\theta_{stu}) &= scr_{o} \cdot \mathcal{L}_{o}(\theta_{stu}) \\
        &+ scr_{s} \cdot \mathcal{L}_{s}(\theta_{stu}) \\
        &+ scr_{e} \cdot \mathcal{L}_{e}(\theta_{stu}), \\
        \mathcal{L}_{cls}(\theta_{stu}) &= -y\log \hat{y}- (1-y) \log (1-\hat{y}), \\
        \mathcal{L}_{all}(\theta_{stu}) &= \mathcal{L}_{cls}(\theta_{stu}) + \alpha \cdot \mathcal{L}_{distill}(\theta_{stu}),
    \end{aligned}
\end{equation}
where $\theta_{stu}$ are parameters of the student model, $y$ and $\hat{y}$ are ground-truth and predicted labels, and $\alpha$ is a weighting factor for the distillation loss.

\section{Experiments}
We compare to 17 methods in three groups to answer the following evaluation questions:
\begin{itemize}
\setlength{\itemindent}{6pt}
\setlength{\itemsep}{1pt}
    \item[\textbf{EQ1} ]Can CAS-FEND outperform other methods for fake news detection?
    \item[\textbf{EQ2} ]Does CAS-FEND still work effectively in real-world scenarios?
    \item[\textbf{EQ3} ]How do three types of knowledge contribute to detection performance?
    \item[\textbf{EQ4} ]How do various hyperparameters affect the performance of CAS-FEND?
\end{itemize}
\subsection{Experimental Setup}
\textbf{Datasets.} We employ two widely used fake news detection datasets for academic research: \textbf{Weibo-21}~\cite{Nan_Weibo21} for Chinese and \textbf{GossipCop}~\cite{fakenewsnet} for English. Each dataset contains news content, user comments, the published time, and the ground-truth label. The Weibo21 dataset ranges from 2010-12-15 to 2021-03-31, and the GossipCop dataset ranges from 2000-07-24 to 2018-12-07. 
For each dataset, we did a train-validation-test set split chronologically to simulate real-world scenarios with a ratio of 4:1:1. 
Besides, we also collect \textbf{Online} data from the ``RuiJianShiYao'' fake news detection system, which handles millions of news pieces every day for additional testing.
The dataset statistics of Weibo21 and GossipCop are shown in \tablename~\ref{tab:dataset}, and the Online data contains 489 fake samples and 5,735 real samples.
\begin{table}[htbp]
  \centering
  \small
  \setlength{\tabcolsep}{4.5pt}  
  \caption{Statistics of Weibo21 and GossipCop datasets.}
    \begin{tabular}{lrrrrrr}
    \toprule
    \multicolumn{1}{c}{\multirow{2}[4]{*}{\textbf{Dataset}}} & \multicolumn{3}{c}{\textbf{Weibo21}} & \multicolumn{3}{c}{\textbf{GossipCop}} \\
    & \multicolumn{1}{c}{Train} & \multicolumn{1}{c}{Val} & \multicolumn{1}{c}{Test} & \multicolumn{1}{c}{Train} & \multicolumn{1}{c}{Val} & \multicolumn{1}{c}{Test} \\
    \midrule
    \#Fake & 2,883 & 540   & 539   & 1,816   & 522   & 844 \\
    \#Real & 2,179 & 702   & 724   & 3,775   & 820   & 483 \\
    Total & 5,062 & 1,242 & 1,263 & 5,591 & 1,342 & 1,327 \\
    \bottomrule
    \end{tabular}%
  \label{tab:dataset}%
\end{table}%

\textbf{Compared Methods.} We categorize compared methods into three groups: content-only, distillation-based, and comment-aware methods. 

Group I consists of five content-only methods, which exploit news content for both training and testing: 
\textbf{(1) LLM$_\text{cnt}$}: A zero-shot method that directly prompts an LLM to make veracity judgments with only news content provided; 
\textbf{(2) BERT}~\cite{devlin-2019-bert}: A pre-trained language model widely used as the text encoder for fake news detection~\cite{ dualemotion, Nan_Weibo21}. In our experiments, we finetune the last layer of the pre-trained BERT with other layers frozen. 
\textbf{(3) BERT-Emo}: A simplified version of DualEmo~\cite{dualemotion} which leverage the semantic feature of news content extracted with BERT~\cite{devlin-2019-bert} as well as the emotion feature of news content for detection. The emotion feature are extracted based on the released code from the original paper.
\textbf{(4) EANN-text}~\cite{eann}: A model that aims to learn event-invariant representations for fake news detection. In our experiments, we use its text-only version.
\textbf{(5) ENDEF}~\cite{endef}: An entity debiasing method that can generalize fake news detectors to future data. BERT~\cite{devlin-2019-bert} is used to encode the news content in our experiments.

Group II contains feature-based KD methods (Gr-oup II-1) and response-based KD methods (Group II-2)~\cite{gou2021_kdsurvey}. 
We fix the trained-over CAS-FEND(tea) as the teacher model with BERT, BERT-Emo, EANN-Text, and ENDEF as the student model respectively, which are denoted as BERT$_\text{stu}$, BERT-Emo$_\text{stu}$, EANN-Text$_\text{stu}$, and ENDEF$_\text{stu}$.
Specifically, \textbf{feature-based KD Methods} is a group of methods that train the student model by aligning the overall feature in the student model to $\boldsymbol{f}^t$ in CAS-FEND(tea); \textbf{response-based KD Methods} is a group of methods that train the student model by distilling the fake-real probability distribution from the decision layer in CAS-FEND(tea) to the one in the student model.
\begin{table*}[t]
    \centering
    \small
    \setlength{\tabcolsep}{4.0pt}
    \caption{Performance of content-only, distillation-based methods, and CAS-FEND(stu) on Weibo21 and Gossipcop datasets. The bold values indicate the best performance. ** ($\rho \leq 0.005$) and * ($\rho \leq 0.001$) indicate paired t-test results of CAS-FEND(stu) v.s. the best-performing method in Groups I and II.}
    \begin{tabular}{lllllllllll}
        \toprule
        \multicolumn{1}{l}{\multirow{2}[2]{*}{\textbf{Model}}} & \multicolumn{5}{c}{\textbf{Weibo21}}   & \multicolumn{5}{c}{\textbf{GossipCop}} \\
         &    \multicolumn{1}{c}{macF1} & \multicolumn{1}{c}{Acc} & \multicolumn{1}{c}{AUC} & \multicolumn{1}{c}{F1$_\text{real}$} & \multicolumn{1}{c}{F1$_\text{fake}$} & \multicolumn{1}{c}{macF1} & \multicolumn{1}{c}{Acc} & \multicolumn{1}{c}{AUC} & \multicolumn{1}{c}{F1$_\text{real}$} & \multicolumn{1}{c}{F1$_\text{fake}$} \\
        \midrule
        \multicolumn{9}{l}{\textit{I. Content-Only Methods}}\\
        LLM$_\text{cnt}$ & 0.6795 & 0.6825 & 0.7119 & 0.6486 & 0.7105 & 0.6029 & 0.6774 & 0.6043 & 0.7750 & 0.4309 \\
     BERT  & 0.7625 & 0.7633  & 0.8439 & 0.7749 & 0.7500  & 0.8073 & 0.8259 & 0.8931 & 0.8670 & 0.7477 \\
    BERT-Emo & 0.7680 & 0.7700 & 0.8490 & 0.7871 & 0.7489 & 0.8212 & 0.8386 & 0.8961 & 0.8768 & 0.7655 \\
    ENDEF & 0.7701 & 0.7717 & 0.8477 & 0.7870 & 0.7532 & 0.8298 & 0.8463 & 0.9002 & 0.8826 & 0.7770 \\
    EANN-text  & 0.7212 & 0.7240 & 0.7986 & 0.7467 & 0.6956 & 0.8179 & 0.8348 & 0.8904 & 0.8733 & 0.7626 \vspace{1mm}\\
    \multicolumn{9}{l}{\textit{II-1. Feature-Based Distillation Methods}} \\
    BERT$_\text{stu}$ & 0.7752 & 0.7761 & 0.8389 & 0.7878 & 0.7726 & 0.8242 & 0.8408 & 0.8909 & 0.8782 & 0.7702 \\
    BERT-Emo$_\text{stu}$ & 0.7677 & 0.7694 & 0.8401 & 0.7855 & 0.7500 & 0.8209 & 0.8386 & 0.8897 & 0.8770 & 0.7648 \\
    ENDEF$_\text{stu}$ & 0.7680 & 0.7699 & 0.8452 & 0.7862 & 0.7594 & 0.8299 & 0.8497 & 0.9001 & 0.8823 & 0.7789 \\
    EANN-text$_\text{stu}$ & 0.7342 & 0.7354 & 0.8125 & 0.7580 & 0.7120 & 0.8200 & 0.8375 & 0.8964 & 0.8756 & 0.7613 \vspace{1mm}\\
    \multicolumn{9}{l}{\textit{II-2. Response-Based Distillation Methods}} \\
    BERT$_\text{stu}$ & 0.7698 & 0.7687 & 0.8500 & 0.7721 & 0.7590 & 0.8099 & 0.8297 & 0.8930 & 0.8680 & 0.7476 \\
    BERT-Emo$_\text{stu}$ & 0.7665 & 0.7676 & 0.8482 & 0.7865 & 0.7488 & 0.8197 & 0.8311 & 0.8923 & 0.8729 & 0.7665 \\
    ENDEF$_\text{stu}$ & 0.7577 & 0.7586 & 0.8341 & 0.7686 & 0.7469 & 0.8181 & 0.8349 & 0.8897 & 0.8734 & 0.7627 \\
    EANN-text$_\text{stu}$ & 0.7089 & 0.7114 & 0.7880 & 0.7266 & 0.6913 & 0.8073 & 0.8231 & 0.8865 & 0.8620 & 0.7525 \vspace{1mm}\\
    \textbf{CAS-FEND(stu)} & \textbf{0.7950$^\text{**}$} &  \textbf{0.7957$^\text{**}$} & \textbf{0.8703$^\text{**}$} & \textbf{0.8072$^\text{**}$} & \textbf{0.7828$^\text{*}$} & \textbf{0.8434$^\text{**}$} & \textbf{0.8598$^\text{*}$} & \textbf{0.9026$^\text{*}$} & \textbf{0.8940$^\text{*}$} & \textbf{0.7928$^\text{**}$} \\
    \bottomrule
    \end{tabular}
    \label{tab:results_stu}
\end{table*}

For Group III, we compare CAS-FEND(tea) with four comment-aware methods: 
\textbf{(1) LLM$_\text{cnt+cmt}$}: A zero-shot method that prompts an LLM to make veracity judgments with content and comments provided; 
\textbf{(2) DualEmo}~\cite{dualemotion}: A framework that enhances fake news detection performance with the help of emotions from news content and user comments, and the emotion gap between them. Here, we exploit BERT~\cite{devlin-2019-bert} as the base detector; 
\textbf{(3) dEFEND}~\cite{Shu_dEFEND}: A model that develops a sentence-com-ment co-attention network for fake news detection; 
\textbf{(4) L-Defense}~\cite{Wang_LDefense}: A method that extracts competing groups of evidence from comments and leverages justifications from LLMs based on the evidence for fake news detection.

\textbf{Implementation Details.} 
We prompt GLM-4 for Weibo21 and GPT-3.5-Turbo (\textit{version 0125}) for GossipCop in veracity judgment. 
The sampling temperature is set to 0.1 for GLM-4 and GPT-3.5-Turbo to get definitive answers (see Prompt 1, in which underlined text is only for LLM$_\text{cnt+cmt}$).
For non-LLM methods, we adopt \textit{bert-base-chinese} for Weibo21/ online data and \textit{bert-base-uncased} for GossipCop to encode news content/comments.

\textbf{Evaluation Metrics.} We adopt the following evaluation metrics: macro F1 score (macF1), accuracy (Acc), and macro F1 score for the real/fake class (F1$_\text{real}$/F1$_\text{fake}$) to report experimental results.
\begin{tcolorbox}[title=Prompt 1:  Prompt for Veracity Judgment, boxrule=0pt, left=1mm, right=1mm, top=1mm, bottom=1mm, fontupper=\small]
    \textbf{System Prompt:} Given the following news piece \dashuline{and the corresponding comments}, predict the veracity of this news piece. \dashuline{The comments are collected from social media users.} If the news piece is more likely to be fake, return 1; otherwise, return 0. Please refrain from providing ambiguous assessments such as undetermined. \\
    \textbf{Context Prompt:} news: [\emph{the given news $o$}]; \dashuline{comments: [\emph{user comments ${c_1, c_2, ...}$}]}. The answer (Arabic numerals) is:
\end{tcolorbox}
\begin{table*}
    \centering
    \small
    \setlength{\tabcolsep}{4.0pt}
    \caption{Performance of comment-aware methods and CAS-FEND(tea) on Weibo21 and Gossipcop datasets. The bold values indicate the best performance. $*$ ($\rho \leq 0.005$) and $**$ ($\rho \leq 0.001$) indicate paired t-test results of CAS-FEND(tea) v.s. the best-performing method in Group III.}
    \begin{tabular}{lllllllllll}
        \toprule
        \multicolumn{1}{l}{\multirow{2}[2]{*}{\textbf{Model}}} & \multicolumn{5}{c}{\textbf{Weibo21}}   & \multicolumn{5}{c}{\textbf{GossipCop}} \\
         &    \multicolumn{1}{c}{macF1} & \multicolumn{1}{c}{Acc} & 
         \multicolumn{1}{c}{AUC} &
         \multicolumn{1}{c}{F1$_\text{real}$} & \multicolumn{1}{c}{F1$_\text{fake}$} & \multicolumn{1}{c}{macF1} & \multicolumn{1}{c}{Acc} & 
         \multicolumn{1}{c}{AUC} &
         \multicolumn{1}{c}{F1$_\text{real}$} & \multicolumn{1}{c}{F1$_\text{fake}$} \\
        \midrule
    \multicolumn{9}{l}{\textit{III. Comment-Aware Methods}}\\
    LLM$_\text{cnt+cmt}$ & 0.7597 & 0.7601 & 0.7824 & 0.7506 & 0.7689 & 0.6360 & 0.6654 & 0.6351 &  0.7394 & 0.5326 \\
    L-Defense & 0.7812 & 0.7815 & 0.8551 & 0.7893 & 0.7730 & 0.8308 & 0.8485 & 0.8964 & 0.8856 & 0.7759 \\
    DualEmo & 0.7834 & 0.7837 & 0.8823 & 0.7987 & 0.7925 & 0.8864 & 0.8802 & 0.9341 & 0.9040 & 0.8620 \\
    dEFEND & 0.8078 & 0.8080 & 0.8901 & 0.8118 & 0.8012 & 0.8971 & 0.8953 & 0.9415 & 0.9132 & 0.8773 \\
    \textbf{CAS-FEND(tea)} & \textbf{0.8181$^\text{*}$} & \textbf{0.8187$^\text{*}$} & 
    \textbf{0.9016$^\text{*}$} &
    \textbf{0.8287$^\text{*}$} & \textbf{0.8074$^\text{*}$} & \textbf{0.9188$^\text{*}$} & \textbf{0.9261$^\text{**}$} & 
    \textbf{0.9716$^\text{**}$} &
    \textbf{0.9432$^\text{**}$} & \textbf{0.8944$^\text{*}$} \\
    \bottomrule
    \vspace{-0.8cm}
    \end{tabular}
    \label{tab:results_tea}
\end{table*}
\vspace{-0.5cm}
\subsection{Fake News Detection Performance (EQ1)}
To evaluate the effectiveness of CAS-FEND(stu) compared to content-only (Group I) /distillation-based methods (Group II) and CAS-FEND(tea) compared to comment-aware methods (Group III), we conduct experiments on Weibo21~\cite{Nan_Weibo21} and GossipCop~\cite{fakenewsnet}.
From the experimental results listed in Tables~\ref{tab:results_stu} and \ref{tab:results_tea}, we have the following findings:

(1) \textbf{CAS-FEND(stu) outperforms content-only methods (Group I).}
Comparing with methods in Group I, we find that CAS-FEND(stu) outperforms all content-only methods, which indicates that CAS-FEND(stu) has obtained additional useful knowledge beyond news content.
Specifically, LLM$_{cnt}$ performs worst for the limited veracity judgment ability of the LLM with news content provided.
For non-LLM methods, BERT suffers from performance degradation in non-independent identical distribution (non-IID) scenarios like temporal distribution shift in our experiments;
BERT-Emo and ENDEF perform better than BERT because the emotional feature is more robust than the semantic feature in different time periods and entity bias mitigating enhances the generalizing ability of fake news detectors; 
EANN learns invariant features thr-ough adversarial training, but it performs differently for Weibo21 and GossipCop. For the GossipCop dataset, EANN brings a few improvements over BERT but has the opposite effect on the Weibo-21 dataset. It may be because the topics of Weibo21 are more varied and lead to more difficulty in extracting invariant features.
Therefore, the improvement of the detection performance is limited only based on news content, and it is necessary to absorb knowledge from external sources beyond news content, such as user comments.

(2) \textbf{CAS-FEND(stu) outperforms distillation methods (Group II).} 
The results in Groups II-1 and II-2 show that only a few models get a performance improvement with feature-based or response-based distillation strategy (BERT and EANN-Text with feature-based KD; BERT with response-based KD) and models in Group II perform inferior to CAS-FEND(stu), which proves the importance and effectiveness of the CAS-FEND(stu) structure and the distillation strategy in CAS-FEND. 
Besides, that the feature-based distillation strategy is more effective than the response-based one is the rationale behind overall knowledge distillation in CAS-FEND. 
\vspace{-0.5cm}

(3) \textbf{CAS-FEND(tea) outperforms other com-ment-aware methods.} 
The results in Group III show that CAS-FEND(tea) outperforms other com-ment-aware methods.
The superior performance of  CAS-FEND(tea) benefits from both semantic and emotional knowledge from user comments, which enhances the understanding of news and improves the detection ability, making it qualified to guide CAS-FEND(stu).
Specifically, CAS-FEND(tea) outperforms dEFEND (the best-performing comment-aware compared methods) by \textbf{1.3\%} macF1 on Wei-bo21 and \textbf{2.4\%} macF1 on GossipCop.

\begin{table*}
\centering
\small
\setlength{\tabcolsep}{5.0pt}
\caption{Macro F1 scores of CAS-FEND(stu) with no comments (0\%), and L-Defense, DualEmo, dEFEND, CAS-FEND(tea) with different percentages of comments available.  For CAS-FEND(tea) and other comment-aware methods, the proportion of comments is set to 25\%, 50\%, 75\%, or 100\%. $\downarrow$ indicates worse performance than CAS-FEND(stu).}
\begin{tabular}{lllllllllll}
  \toprule
  \multicolumn{1}{l}{\multirow{2}[2]{*}{\textbf{Model}}} & \multicolumn{5}{c}{\textbf{Weibo21}}   & \multicolumn{5}{c}{\textbf{GossipCop}} \\
   &    \multicolumn{1}{c}{0\%} & \multicolumn{1}{c}{25\%} & \multicolumn{1}{c}{50\%} & \multicolumn{1}{c}{75\%} & \multicolumn{1}{c}{100\%} & \multicolumn{1}{c}{0\%} & \multicolumn{1}{c}{25\%} & \multicolumn{1}{c}{50\%} & \multicolumn{1}{c}{75\%} & \multicolumn{1}{c}{100\%} \\
  \midrule
  CAS-FEND(stu) & \textbf{0.7950} & \multicolumn{1}{c}{-} & \multicolumn{1}{c}{-} & \multicolumn{1}{c}{-} & \multicolumn{1}{c}{-} & \textbf{0.8434} & \multicolumn{1}{c}{-} & \multicolumn{1}{c}{-} & \multicolumn{1}{c}{-} & \multicolumn{1}{c}{-} \\
  L-Defense & \multicolumn{1}{c}{-} & 0.7671$\downarrow$ & 0.7727$\downarrow$ & 0.7744$\downarrow$ & 0.7812$\downarrow$ & \multicolumn{1}{c}{-} & 0.8024$\downarrow$ & 0.8068$\downarrow$ & 0.8138$\downarrow$ & 0.8308$\downarrow$ \\
  DualEmo & \multicolumn{1}{c}{-} & 0.7501$\downarrow$ & 0.7621$\downarrow$ & 0.7753$\downarrow$ & 0.7834$\downarrow$ & \multicolumn{1}{c}{-} & 0.8249$\downarrow$ & 0.8447 & 0.8776 & 0.8864 \\
  dEFEND & \multicolumn{1}{c}{-} & 0.7944$\downarrow$ & 0.8014 & 0.8107 & 0.8125 & \multicolumn{1}{c}{-} & 0.8344$\downarrow$ & 0.8541 & 0.8907 & 0.8971 \\
  CAS-FEND(tea) & \multicolumn{1}{c}{-} & 0.7941$\downarrow$ & \textbf{0.8015} & \textbf{0.8112} & \textbf{0.8181} & \multicolumn{1}{c}{-} & 0.8421$\downarrow$ & \textbf{0.8713} & \textbf{0.9112} & \textbf{0.9188} \\
  \bottomrule
\end{tabular}
\label{tab: comment_num_test}
\end{table*}
\begin{table}[htbp]
  \centering
  \small
  \setlength{\tabcolsep}{2.8pt}
  \caption{Relative performance improvement of three content-only models and CAS-FEND(stu) over BERT on the online dataset.}
    \begin{tabular}{lrrrr}
    \toprule
          \multicolumn{1}{c}{\textbf{Model}} & 
          \multicolumn{1}{c}{SPAUC} & \multicolumn{1}{c}{AUC} & \multicolumn{1}{c}{macF1} & \multicolumn{1}{c}{Acc}\\
    \midrule
    \multicolumn{5}{l}{\textit{I. Content-Only Methods}} \\
    BERT-Emo & -0.0205 & -0.0040 & +0.0071 &  +0.0113\\
    ENDEF & +0.0076 & +0.0227 & +0.0139 & +0.0277\\
    EANN-text  & -0.0684 & -0.0777 & -0.1575 & -0.0761\\
    \multicolumn{5}{l}{\textit{II-1. Feature-Based Distillation Methods}} \\
    BERT$_\text{stu}$ & 0.0016 & 0.0157 & 0.0094 & 0.0191  \\
    BERT-Emo$_\text{stu}$ &  -0.0194 & 0.0008 & 0.0092 & 0.0199  \\
    ENDEF$_\text{stu}$ & 0.0135 & 0.0272 & 0.0217 & 0.0288  \\
    EANN-text$_\text{stu}$ & -0.0567 & -0.0686 & -0.0503 & -0.0700  \\
    \multicolumn{5}{l}{\textit{II-2. Response-Based Distillation Methods}} \\
    BERT$_\text{stu}$ & 0.0002 & 0.0038 & 0.0035 & 0.0131 \\
    BERT-Emo$_\text{stu}$ &  -0.0293 & -0.0172 & 0.0024 & 0.0085  \\
    ENDEF$_\text{stu}$ & -0.0089 & 0.0133 & -0.0088 & 0  \\
    EANN-text$_\text{stu}$ & -0.069 & -0.0857 & -0.0568 & -0.0831 \\
    CAS-FEND(stu)  & \textbf{+0.0314} & \textbf{+0.0243} & \textbf{+0.0388} & \textbf{+0.0401} \\
    \bottomrule
    \vspace{-0.5cm}
    \end{tabular}%
  \label{tab: online}%
\end{table}%
\subsection{Applicability in Real-World Scenarios (EQ2)}
We evaluate the applicability of CAS-FEND in real-world scenarios from two aspects: 
One is a dynamic scene with a growing number of comments, and another is an unbalanced scenario with the number of real news exceeds that of fake news by far.

\paragraph{\textbf{Scenario I: A Growing Number of Comments}}
Since a news article is created and published, there will be a growing number of user comments over time. 
It is more challenging to maintain the detection performance in the early stage and after a period of dissemination simultaneously.
To investigate the corresponding ability of CAS-FEND, it is worthwhile to explore whether CAS-FEND possesses a sustainable competitive performance with various numbers of comments, \textit{i.e.}, from less to more.
We simulate this real-world scenario by sampling different subsets of comments.
Specifically, for each news piece in the testing set, we arrange the comments chronologically and sample 0\%, 25\%, 50\%, 75\%, and 100\% comments from front to back to get the corresponding subsets. 
The performance of CAS-FEND(stu)/content-only methods/KD methods with 0\% comments and comment-aware methods with 100\% comments are exactly listed in Tables~\ref{tab:results_stu} and \ref{tab:results_tea}.
We additionally test comment-aware methods with 25\%, 50\%, and 75\% comments, and show the performance in Table~\ref{tab: comment_num_test}\footnote{We do not list the performance of content-only, KD methods because CAS-FEND(stu) outperforms all of them and can serve as a representative with 0\% comments.}.

From the results, we find that: (1) CAS-FEND(tea) performs best with 50\%-100\% comments, indicating CAS-FEND(tea) makes the best use of comments; (2) CAS-FEND(stu) outperforms comment-aware methods with 25\% comments for both datasets and even DualEmo with 100\% comments on Weibo21, which demonstrates that CAS-FEND(stu) has absorbed beneficial knowledge of user comments from the teacher model and improved the early detection capability.
In conclusion, CAS-FEND possesses the best performance by switching between CAS-FEND(stu) and CAS-FEND(tea) according to the proportion of user comments (Specifically, choose the student model for 0\%-25\% comments and the teacher model for 50\%-75\% comments).
\begin{table*}[htbp]
  \centering
  \small
  \caption{Ablation study on the Weibo21 and the GossipCop datasets.}
    \begin{tabular}{lcccccccccc}
    \toprule
    \multicolumn{1}{l}{\multirow{2}[2]{*}{\textbf{Method}}} & \multicolumn{5}{c}{\textbf{Weibo21}}   & \multicolumn{5}{c}{\textbf{GossipCop}} \\
     & \multicolumn{1}{c}{macF1} & \multicolumn{1}{c}{Acc} & \multicolumn{1}{c}{AUC} & \multicolumn{1}{c}{F1$_\text{real}$} & \multicolumn{1}{c}{F1$_\text{fake}$} & \multicolumn{1}{c}{macF1} & \multicolumn{1}{c}{Acc} & \multicolumn{1}{c}{AUC} & \multicolumn{1}{c}{F1$_\text{real}$} & \multicolumn{1}{c}{F1$_\text{fake}$} \\
    \midrule
    \textbf{CAS-FEND(stu)} & \textbf{0.7950} & \textbf{0.7957} & \textbf{0.8703} & \textbf{0.8072} & \textbf{0.7828} & \textbf{0.8434} & \textbf{0.8598} & \textbf{0.9026} & \textbf{0.8940} & \textbf{0.7928}\\
    \quad$w/o$ semantic & 0.7820 & 0.7824 & 0.8626 & 0.7906 & 0.7735 & 0.8255 & 0.8396 & 0.8986 & 0.8749 & 0.7761\\
    \quad$w/o$ emotional & 0.7797 & 0.7809 & 0.8547 & 0.7952 & 0.7643 & 0.8264 & 0.8405 & 0.8976 & 0.8755 & 0.7774\\
    \quad$w/o$ overall & 0.7785 & 0.7798 & 0.8553 & 0.7947 & 0.7602 & 0.8241 & 0.8385 & 0.8954 & 0.8728 & 0.7742\\
    \bottomrule
    \end{tabular}%
  \label{tab:ablation}%
\end{table*}%
\paragraph{\textbf{Scenario II: Unbalanced Fake-Real Categories}} 
It was known that a gap exists between real-world practicality and the claimed performance of some existing detection models~\cite{xiao2023challenges}. 
Therefore, we test CAS-FEND(stu) and other content-only methods on the online dataset that are newly emerging news (no comments accumulated) with highly skewed real-fake distribution (real: fake $\approx$ 12: 1), which fits the real-world scenario well and also brings great challenges to detectors. 
In real-world scenarios, we should detect fake news without misclassifying real news as much as possible. In other words, we aim to improve the True Positive Rate (TPR) while lowering the False Positive Rate (FPR). Therefore, we additionally adopt Standardized Partial AUC (SPAUC$_{\text{FPR} \leq 0.1}$)~\cite{Donna_Medical1989_SPAUC, Zhu_M3FEND}, which evaluates the area under the ROC curve when FPR does not exceed a specific value (here, 0.1). Due to company regulations, we only report the \textbf{relative improvements} over the BERT model. 
The experimental results in \tablename~\ref{tab: online} show that CAS-FEND(stu) performs best for the real system, indicating that it is robust to the ratio of fake/real categories and applicable to unbalanced scenarios.

\subsection{Analysis of Three Types of Knowledge (EQ3)}
\subsubsection{Ablation Study}
To analyze the effects of the three types of knowledge in CAS-FEND, we conduct an ablation study with three types of variant models \emph{w/o} sem, \emph{w/o} emo, and \emph{w/o} overall, which remove semantic-level, emotional-level, and overall knowledge distillation respectively. 
Accordingly, we adjust the Knowledge Preference Scorer to output two scores to guarantee that the remaining two types of knowledge can be weighted adaptively. 
Experimental results are shown in \tablename~\ref{tab:ablation}.
Comparing the performance of CAS-FEND(stu) with three variant models \emph{w/o} sem, \emph{w/o} emo, and \emph{w/o} overall, we find that removing any type of knowledge causes performance degradation of the student model, which proves the importance of all three types of knowledge.
\begin{table*}[htbp]
    \centering
    \small
    \caption{Two representative examples with the highest emotional and semantic scores respectively.}
    \begin{tabular}{lp{6.8cm}p{6.8cm}}
    \toprule
    \multicolumn{1}{c}{\bm{$scr_e: scr_s: scr_o$}} & \multicolumn{1}{c}{\textbf{News Content}} & \multicolumn{1}{c}{\textbf{Key Infomation in User Comments}}\\
    \midrule
    \textbf{0.65}: 0.25: 0.10 & According to BBC, actor Tom Cruise has died after his inflatable boat exploded while filming. & What?!; ??? ;  crazy?; Unbelievable!!; ??? Impossible; Don't talk nonsense!!! \vspace{2.0mm}\\ 
    0.13: \textbf{0.57}: 0.30 & Maserati female owner's drunk driving caused two deaths. There is now a saying that the female owner has intermittent \underline{mental illness}. Really do not want this woman to reduce the sentence or light sentence. & This is the first time I've heard of a connection between drunk driving and \underline{psychosis}; Using \underline{mental illness} as a shield again; Why can \underline{mental illness} get a driver's license? \\
    \bottomrule
    \end{tabular}
    \label{tab:cases}
\end{table*}
\begin{figure}[htbp]
\centering
\includegraphics[width=1.0\linewidth]{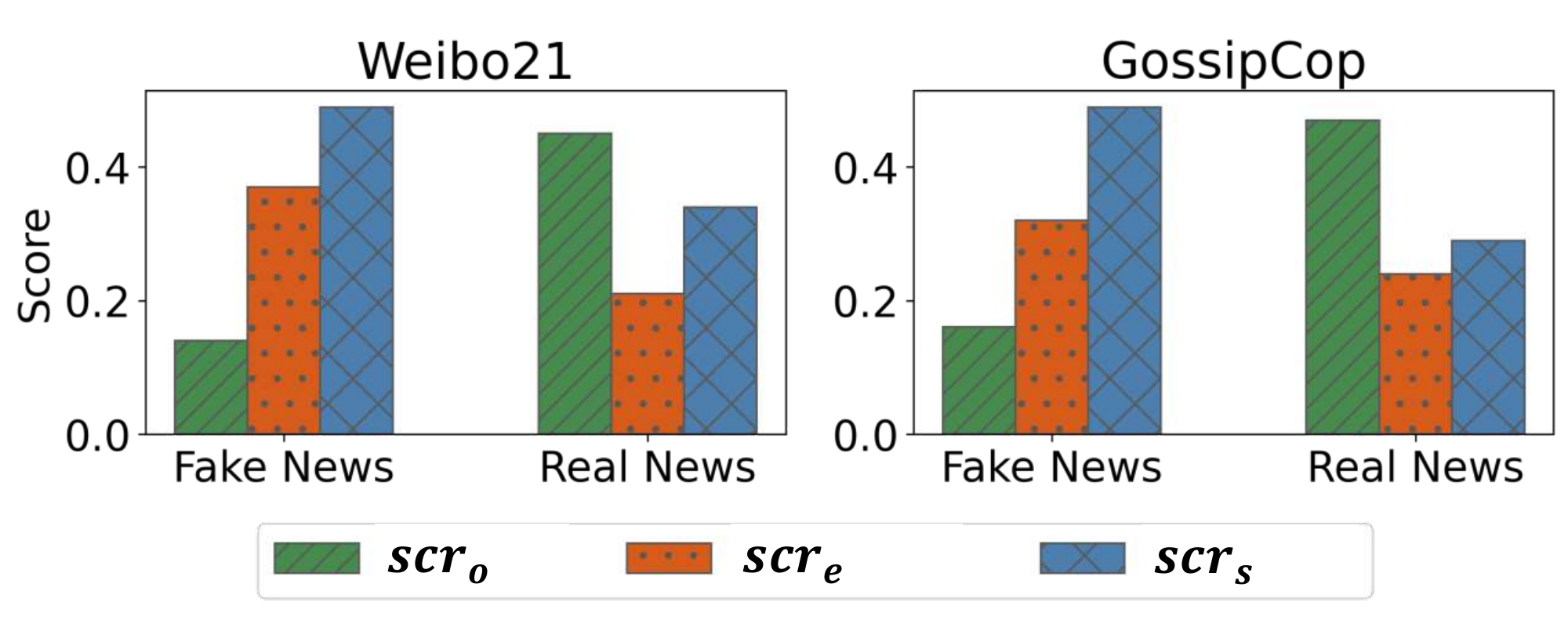}
\caption{Distribution of average preference scores for fake and real categories.}
\vspace{-0.4cm}
\label{fig:relative_score}
\end{figure} 
\subsubsection{Relative Importance Analysis} 
\label{sec: relative_importance_average}
To explore the relative importance of semantic, emotional, and overall knowledge, we calculate the average scores of three types of knowledge respectively. 
\figurename~\ref{fig:relative_score} shows the score distribution on fake and real categories.
From the bar chart, it can be observed that: 
(1) emotional and semantic knowledge play a more significant role in fake news than in real news for both datasets; 
(2) the preference score of emotional knowledge is higher than that of semantic-level knowledge for fake news, which is consistent with existing research that fake news arouses more intense emotion than real news. 
\vspace{-0.5cm}
\subsubsection{Case Analysis}
Based on the findings in Section~\ref{sec: relative_importance_average} that emotional- and semantic-level knowledge contributes greatly to fake news, we conduct a case analysis for fake instances. 
We analyze two cases from the testing set with the highest preference score on emotional- and semantic-level knowledge respectively, and the representative examples are shown in \tablename~\ref{tab:cases}.
From the news content and key information in user comments, we find that emotional-level knowledge contributes the most when user comments consist of intense emotions, and semantic-level knowledge he-lps to capture the interlinked information between news content and user comments (\underline{underlined} content in the second example).
\begin{figure}[ht]
\centering
\includegraphics[width=1.0\linewidth]{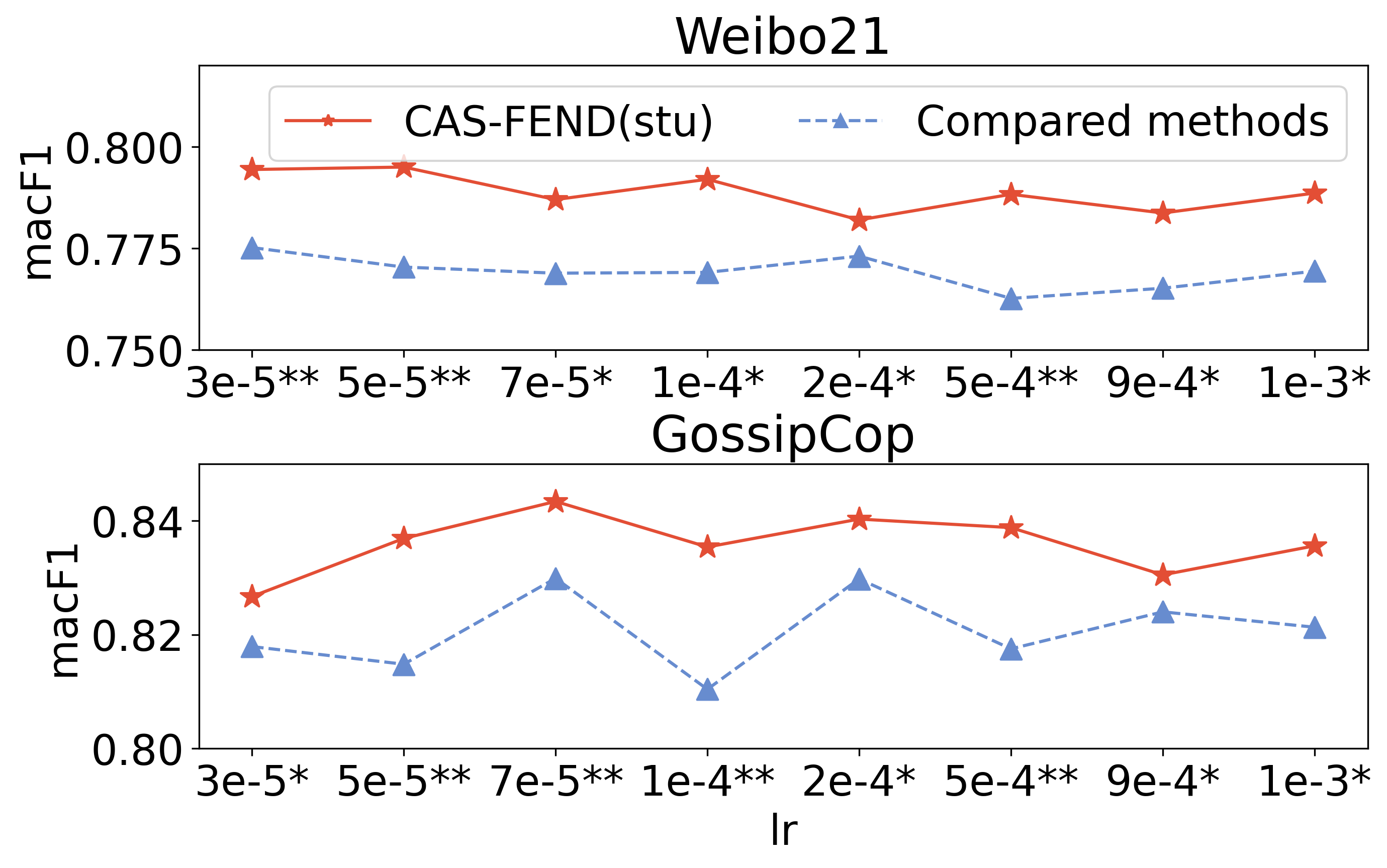}
\caption{Macro F1 scores of CAS-FEND(stu) (solid red lines) and the best-performing method of compared methods in Groups I \& II (dotted blue).  * ($\rho \leq 0.005$) and ** ($\rho \leq 0.001$) indicate paired t-test of CAS-FEND(stu) v.s. the best-performing method in Groups I \& II.}\label{fig:lr_sensitivity}
\vspace{-0.6cm}
\end{figure} 
\subsection{Impact of Hyperparameters (EQ4)}
We explore the impact of the initial learning rate, the weighting factor for distillation, and the number of comments in the training stage on the performance of CAS-FEND(stu).

\begin{figure}[ht]
\centering
\includegraphics[width=1.0\linewidth]{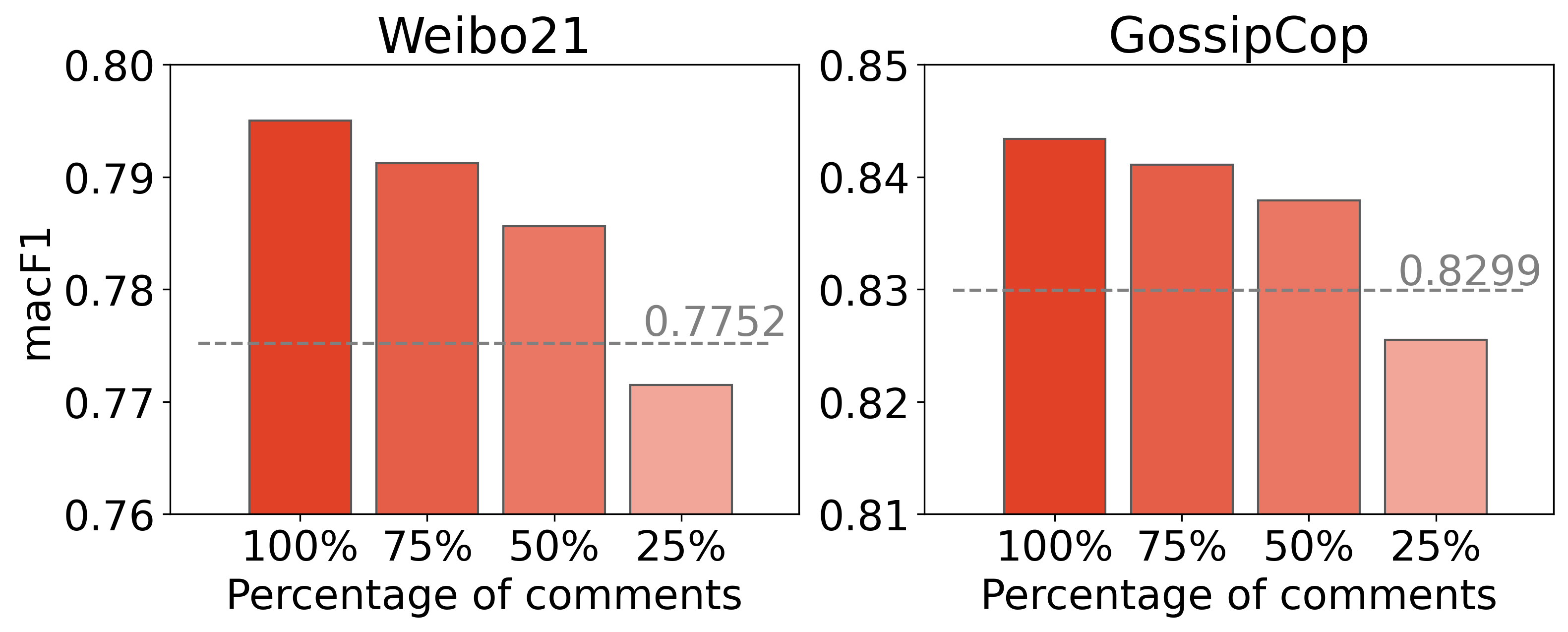}
\caption{Macro F1 scores of CAS-FEND(stu) (solid red bars) and the best-performing method of compared methods in Groups I \& II (dotted gray).}\label{fig:comment_percent}
\vspace{-0.2cm}
\end{figure}

\paragraph{\textbf{Initial Learning Rate}}
To explore the impact of the initial learning rate of the student model training stage on the performance of CAS-FEND(stu), we conduct experiments by training the student mo-del with initial learning rates of $3e-5, 5e-5, 7e-5, 1e-4, 2e-4, 5e-4, 9e-4$, and $1e-3$. 
The performance of CAS-FEND(stu) and the best-performing method in Groups I and II are shown in \figurename~\ref{fig:lr_sensitivity}. 
Each data point is calculated by averaging 10 independent repeated experimental results.
The comparison indicates that CAS-FEND(stu) consistently outperforms other methods with different learning rates, and $5e-5, 7e-5$ are the best learning rates for Weibo21 and GossiopCop respectively.
\begin{figure}[ht]
\includegraphics[width=1.0\linewidth]{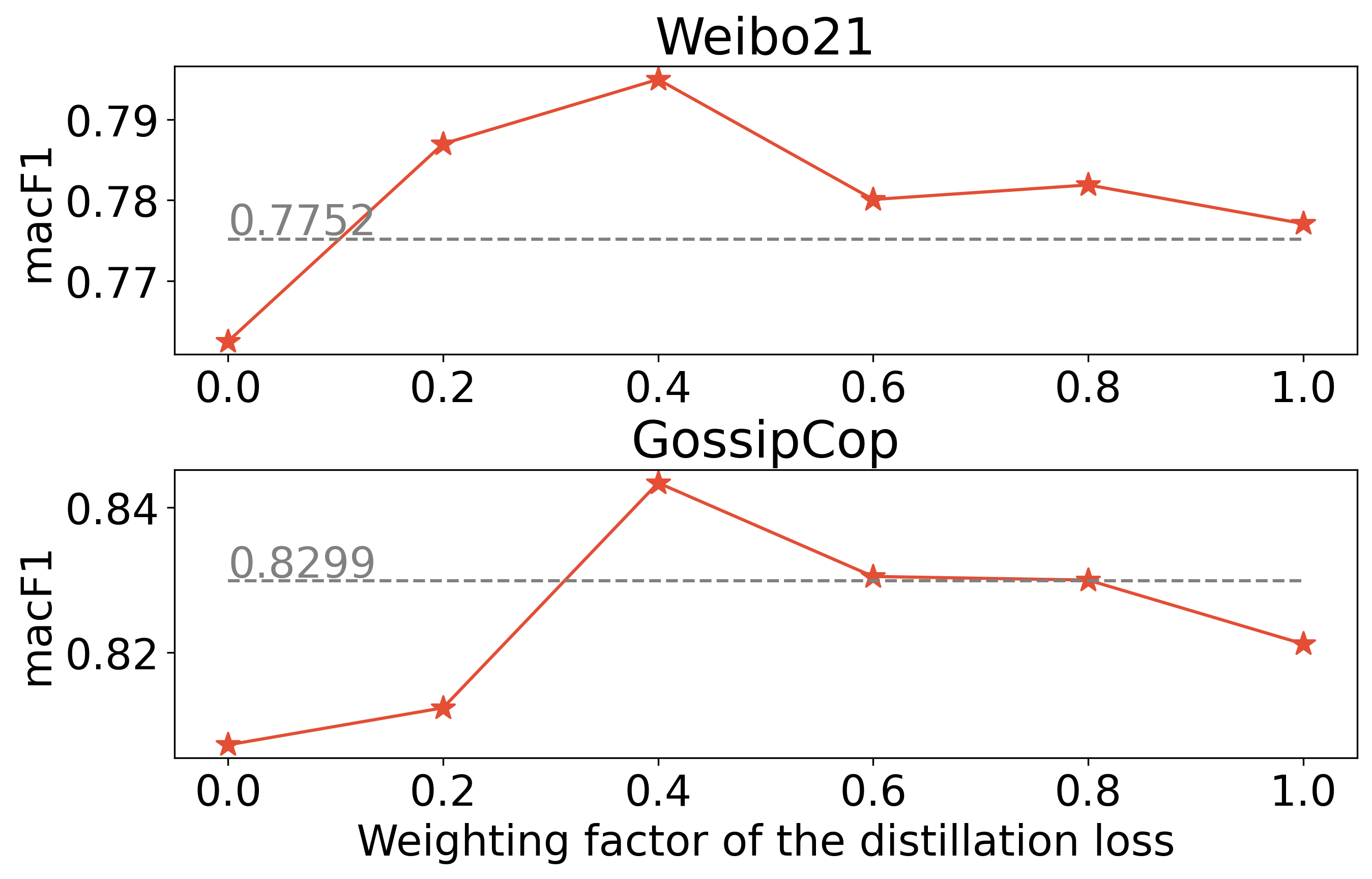}
\caption{Macro F1 scores of CAS-FEND(stu) (solid red lines) and the best-performing method of compared methods in Groups I \& II (dotted gray).}\label{fig:weighting_factor}
\vspace{-0.4cm}
\end{figure}
\paragraph{\textbf{Weighting Factor for Distillation}}
The weighting factor $\alpha$ controls the contribution of the distilled knowledge. 
To explore the impact of the weighting factor on CAS-FEND(stu)'s performance, we conduct experiments with different weighting factors as $0.0, 0.2, 0.4, 0.6, 0.8$, and $1.0$ respectively, where $0.0$ represents distilling no knowledge from the teacher model.
From experimental results shown in \figurename~\ref{fig:weighting_factor}, we find that $0.4$ achieves the best performance for both datasets, which indicates that a proper weight for distilling knowledge( not too low or too high) is best.
\paragraph{\textbf{Proportion of Available Comments of Each\\ Training Data}}
In the default setting, we exploit all available comments (100\%) during the teacher model training procedure, which are listed in \tablename~\ref{tab:results_tea}. 
Here we investigate how our model performs when the number of comments in each training sample decreases to relatively small scales, corresponding to the proportion of 75\%, 50\%, and 25\% with the earlier posted comments preserved as priority. 
From the results shown in \figurename~\ref{fig:comment_percent}, we find that: (1) Aligned with our intuition, the macro F1 decreases as the number of available comments for training decreases because of the loss of crowd feedback information; 
(2) With at least half of the comments preserved, CAS-FEND(stu) can outperform the best-performing method, showing the effectiveness of the earlier comments and the success of knowledge transfer. 
\section{Discussion}
In this section, we make a deeper discussion about our proposed method from the following aspects:

\textbf{Meet the requirements of real systems.} The characteristics of social media, such as a large number of users, massive information volume, and fast dissemination speed, lead to the rapid reach of fake news to a large number of users shortly after its release. CAS-FEND, which improves accuracy and simultaneously guarantees timeliness, is expected to serve as a useful tool to combat fake news as early as possible. 
CAS-FEND leverages additional information from user comments in the training procedure to enhance the inference ability with only news content.
Therefore, CAS-FEND has a low deployment cost with better performance compared to other content-only methods, which is advantageous for real-world systems. 

\textbf{Mitigate the temporal distribution shift.} Many researchers~\cite{Mu_ACL2023_Rethinking, hu-etal-2023-learn} demonstrate the existence of temporal distribution shifts between newly emerging news and historical news. Our temporal dataset split corresponds to this situation and could reflect the capability of detection methods to adapt to future data. In such a challenging setting, our proposed CAS-FEND(stu) still outperforms a series of compared methods, indicating that CAS-FEND(stu) could better mitigate the temporal distribution shift in news data. 
We attribute the superiority of CAS-FEND(stu) to the effective introduction of comment-aware signals via distillation from the teacher model. Compared to news content, collective feedback like comments generally provides more stable signals across different time periods. 

\textbf{Limitation and future work.} (1) A well-perfor-ming teacher model is an essential precondition for effectively training a well-performing student model. Therefore, CAS-FEND may be less effective when all of the available comments are of low quality or quantity. It is worth exploring the use of large language models as an analyzer or data generator~\cite{hu2024bad,Nan_GenFEND, SIGIR_tutorial};
(2) The process of distillation requires additional computational overhead during training, wh-ich may be unfriendly to cost-sensitive developers;
(3) It is promising to investigate how to effectively exploit other types of social context information for early fake news detection.
\section{Conclusion}
In this paper, we investigated the problem of guaranteeing accuracy and timeliness for fake news detection by making up for the lack of user comments for newly emerging news.
We proposed to find a surrogate one by absorbing comment knowledge from historical news.
As an instantiation, we designed a Comments Assisted Fake News Detection method (CAS-FEND), which adopts a teacher-student architecture and aims to absorb the necessary knowledge of user comments from the comm-ent-aware teacher model and inject it into the con-tent-only student model.
From the semantic perspective, we performed co-attention operation between news content and user comments to extract useful semantic features from both sides.
From the emotional perspective, we extracted the social emotion feature from user comments which is arou-sed by the corresponding news.
We aggregated semantic features of news content/user comments and the social emotion feature to train the teacher model.
To get a well-performing content-only student model, we distilled knowledge from the frozen well-trained teacher model to the student model from semantic, emotional, and overall perspectives adaptively.
Comprehensive experiments on two public datasets (Weibo21 and GossipCop) and the online test demonstrated the superiority and effectiveness of both CAS-FEND(stu) and CAS-FEND(tea).

\bibliographystyle{sty}
\bibliography{ref}

\end{document}